\documentclass[10pt, conference, letterpaper]{IEEEtran}
\usepackage[utf8]{inputenc}
\usepackage{listings}
\usepackage{amsmath, amssymb, amsfonts, amsthm}
\usepackage{mathtools}
\usepackage[nolist,nohyperlinks]{acronym}
\usepackage{float} 
\usepackage[table,xcdraw]{xcolor}
\usepackage[most]{tcolorbox}
\usepackage[normalem]{ulem}
\usepackage{subcaption}
\usepackage{hyperref}
\usepackage{color, soul}
\usepackage{booktabs}
\usepackage{multirow}
\usepackage{multicol}
\usepackage{tabularray}
\usepackage{bbm}
\DeclareUnicodeCharacter{2061}{}
\usepackage[top=0.72in, bottom=1.0in, left=0.63in, right=0.66in]{geometry} 
\usetikzlibrary{positioning, shapes.multipart, arrows.meta}

\def\BibTeX{{\rm B\kern-.05em{\sc i\kern-.025em b}\kern-.08em T\kern-.1667em\lower.7ex\hbox{E}\kern-.125emX}}

\def\vectb#1{\mathbf{#1}}
\def\cal#1{\mathcal{#1}}

\newcolumntype{z}{>{\centering\arraybackslash}m{2.5cm}}
\newcolumntype{q}{>{\columncolor{black!10}}p}

\makeatletter
\newcommand{\substackleft}[1]{%
  \vcenter{%
    \Let@ \restore@math@cr \default@tag
    \baselineskip\fontdimen10 \scriptfont\tw@
    \advance\baselineskip\fontdimen12 \scriptfont\tw@
    \lineskip\thr@@\fontdimen8 \scriptfont\thr@@
    \lineskiplimit\lineskip
    \ialign{$\m@th\scriptstyle##$&$\m@th\scriptstyle{}##$\hfil\crcr
      #1\crcr
    }%
  }%
}
\makeatother

\author{\IEEEauthorblockN{Mohamed Sana}\\
\IEEEauthorblockA{Huawei, France\\
Email: mohamed.sana@huawei.com}}

\title{Reasoning Language Models for Root Cause Analysis in 5G Wireless Networks}

\author{
Mohamed Sana$^{\dagger}$, Nicola Piovesan$^{\dagger}$, Antonio De Domenico$^{\dagger}$, Yibin Kang$^*$,\\ Haozhe Zhang$^*$, Merouane Debbah$^\ddagger$, Fadhel Ayed$^{\dagger}$\\
$^{\dagger}$Paris Research Center, Huawei Technologies, Boulogne-Billancourt, France\\
$^*${Huawei Technologies, China}\\
$^\ddagger${Khalifa University of Science and Technology, Abu Dhabi, UAE}\\
}

\date{May 2025}

\begin{document}

\maketitle

\begin{abstract}
    Root Cause Analysis (RCA) in mobile networks remains a challenging task due to the need for interpretability, domain expertise, and causal reasoning. In this work, we propose a lightweight framework that leverages Large Language Models (LLMs) for RCA.
    To do so, we introduce TeleLogs\footnote{{https://huggingface.co/datasets/netop/TeleLogs}}, a curated dataset of annotated troubleshooting problems designed to benchmark RCA capabilities. 
    Our evaluation reveals that existing open-source reasoning LLMs struggle with these problems, underscoring the need for domain-specific adaptation. 
    To address this issue, we propose a two-stage training methodology that combines supervised fine-tuning with reinforcement learning to improve the accuracy and reasoning quality of LLMs. The proposed approach fine-tunes a series of RCA models to integrate domain knowledge and generate structured, multi-step diagnostic explanations, improving both interpretability and effectiveness. 
    Extensive experiments across multiple LLM sizes show significant performance gains over state-of-the-art reasoning and non-reasoning models, including strong generalization to randomized test variants. These results demonstrate the promise of domain-adapted, reasoning-enhanced LLMs for practical and explainable RCA in network operation and management.
\end{abstract}

\section{Introduction}
Modern mobile networks are complex, large-scale systems that must maintain high level of performance and reliability. Despite extensive monitoring and control mechanisms, faults inevitably arise, ranging from hardware failures to software miss-configurations. While fault detection highlights symptoms of network issues, effective resolution demands a deeper understanding of the underlying causes driving these symptoms.
This is the objective of \ac{RCA}, a critical component of network \ac{OandM} aimed at uncovering the fundamental reasons behind observed problems in the network. Beyond merely pinpointing the source of a fault, \ac{RCA} also seeks to provide comprehensive, factual explanations grounded in system behaviors and domain knowledge to assist engineers in decision making and remediation.
Traditionally, \ac{RCA} models are constructed by integrating expert knowledge into logical frameworks that define causal relationships between observed symptoms and potential root causes.

However, manually encoding these rules and heuristics relies heavily on domain expertise~\cite{zhang2022icassp} and becomes increasingly prohibitive with the growing scale, complexity, and heterogeneity of modern mobile networks.

To address these challenges, machine learning techniques such as decision trees, support vector machines, neural networks, and bayesian networks, have been applied to automate \ac{RCA}~\cite{zhang2022icassp, tong2021machine, gonzalez2017root}. Although these approaches have shown promise, they often face limitations related to scalability, interpretability, and generalization, particularly when dealing with multiple symptoms simultaneously or with high-dimensional data.

Recent progress in \acp{LLM} has opened new opportunities for the design of more advanced RCA models. Their ability to process unstructured data, synthesize domain knowledge, and generate human-readable explanations makes them well suited to network troubleshooting tasks. However, LLMs also present important limitations. Although their outputs are often contextually rich, they frequently lack the formal rigor, consistency and precision required for decision making; qualities typically ensured by rule-based systems.

To overcome these limitations, we propose leveraging \emph{reasoning} \acp{LLM}, which are models specifically fine-tuned for structured, multi-step reasoning. Unlike traditional \acp{LLM}, these models bring key advantages to \ac{RCA}. They are capable of producing coherent diagnostic explanations that combine learned patterns with domain-specific rules, improving both interpretability and practical usability.

This paper presents a \emph{novel} approach to fine-tune reasoning LLMs for \ac{RCA} in 5G mobile networks. We demonstrate how domain knowledge can be systematically integrated into the reasoning process of LLMs to enhance both the accuracy and interpretability of fault diagnosis.

The contributions of the paper are as follows.

\begin{itemize}
    \item We introduce the TeleLogs, a curated dataset of network troubleshooting scenarios with expert-level annotations. TeleLogs comprises both training and testing data and is publicly released to support research in RCA.  
    \item We evaluate open-source LLMs on TeleLogs and show that even state-of-the-art reasoning models struggle to solve the troubleshooting tasks, highlighting their complexity and the limitations of existing LLMs in this domain.
    \item Motivated by these findings, we design a novel two-stage training methodology. The first stage involves supervised fine-tuning  using an LLM-driven multi-agent pipeline that generates diverse and structured \ac{CoT} traces, embedding domain knowledge into the reasoning process. In the second stage, we apply reinforcement learning using \ac{GRPO} to further improve the diagnostic performance and reasoning ability of the model.
\end{itemize}

\section{Related work}

In network \ac{OandM}, practical \ac{RCA} frameworks have traditionally relied on fault trees, which identify root causes based on rules defined manually by expert engineers \cite{sole2017survey}. However, these approaches suffer from several limitations. They lack generalizability, as each new fault scenario often requires the development of a dedicated tree, and they struggle to capture complex causal relationships. Moreover, they are constrained by the scope and availability of expert knowledge, which limits their scalability.
To overcome these limitations, graph-based methods have been proposed as a mean to model dependencies among network components more effectively.

For example, \cite{Yen22GNN} combines \acp{GNN} with graph structure learning to model interdependencies in multivariate time-series data, enabling root cause inference from the learned graph.

Similarly, \cite{mata24SpatialGraph} employs spatial graph convolutional networks to model the spatial and physical dependencies among \acp{BS}, enabling the prediction of performance degradation at a target \ac{BS}. While these methods show promise, they are generally limited to classification tasks and often focus on a single \ac{BS}. Furthermore, although some methods incorporate mechanisms to explain the identified root cause, they fall short in generating comprehensive, human-readable rationales grounded in system behavior and domain knowledge.

Recently, LLM-based approaches have emerged as promising alternatives to graph-based RCA. In \cite{tnAutoRCA2025}, the authors propose Auto-RCA an agentic system of LLMs for alarm-based RCA, which iteratively refines a code-based solution to evaluate, analyze, and repair systematic failures that raise alarms in the network. In \cite{roy2024exploring}, the authors leverage the ReAct paradigm, i.e., LLMs capable of interleaving reasoning steps with the invocation of external tools, to diagnose cloud infrastructure incidents by integrating LLMs with external information retrieval and access to diagnostic modules. Similarly, \cite{wang2024rcagent} proposed RCAgent, a tool-enhanced LLM system deployed in private cloud environments, which extends ReAct with self-consistency, wherein the final answer is derived by aggregating multiple reasoning trajectories produced by the agent. 
More recently, \cite{pei2025flow} introduced Flow-of-Action, an LLM-driven multi-agent framework that explicitly guides the RCA process by structuring the diagnosis into step-by-step summaries aligned with standard operating procedures.

Although LLMs offer a new paradigm for RCA by combining data-driven reasoning with natural language explanations, they still face limitations in terms of formal interpretability. In particular, there remains a gap in integrating the explainability of rule-based systems with the reasoning capabilities of LLMs.

\begin{figure}
    \centering
    \includegraphics[width=\linewidth]{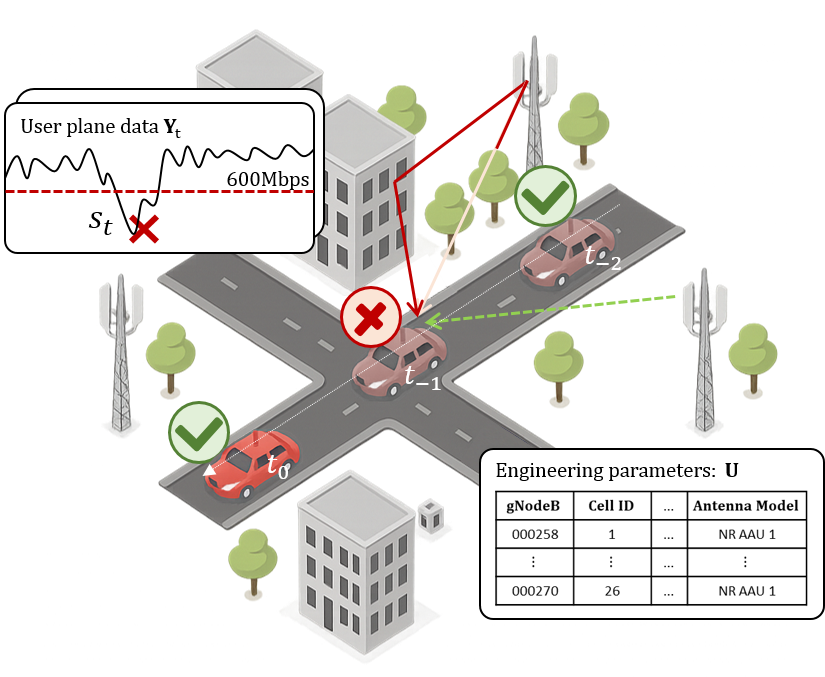}
    \caption{5G Network drive testing.}
    \label{fig:drive-testing}
\end{figure}

\section{Problem Overview}

In this section, we formally define the \ac{RCA} problem in the context of mobile networks. We describe the key {network state} variables and observed symptom, and formulate RCA as a probabilistic inference task.

\subsection{Problem formulation}

Given a set of data describing the current state of the network, our objective is twofold: i) to identify the most probable cause that best explains the observed symptom from a predefined set of potential root causes; and ii) to generate a structured, step-by-step explanation supporting this diagnosis. 
This problem is described by the state of the network represented as a combination of network engineering parameters and observations collected by network drive testing (see Figure \ref{fig:drive-testing}), the symptoms inferred in the observations, and a predefined set of possible root causes.\\[0.5em] 
\noindent
\textbf{Network engineering parameters.} Let $\vectb{U} = \{\vectb{u}_1, \vectb{u}_2, \dots, \vectb{u}_P\}$ be a set of $P$ configuration parameters, where each $\vectb{u}_i\in\cal{U}$ is a vector of categorical and/or numerical features. In this work, as shown in Table \ref{tab:config-data}, $\vectb{u}_i$ represents the configuration parameters of cell $i$, including, cell ID, number of TX and RX antennas, beamforming scenario.\\[0.5em]
\noindent
\textbf{Observations.} Let $\vectb{Y}_t = \{\vectb{y}_{1, :t}, \vectb{y}_{2, :t}, \dots, \vectb{y}_{M, :t}\}$ be a set of $M$ categorical or numerical time-series data, which represents measurements of the network collected through drive testing. {In this work, this observations represent to the user-plane data as shown in Table \ref{tab:user-plane-data}.}\\[0.5em] 
\noindent
\textbf{Symptoms.} Let $s_t\in\mathcal{S}$ be a symptom\footnote{Although many symptoms can be observed in the network, we focus here on the case of a single symptom.} observed in the network, which may be inferred from the measurement data $\vectb{Y}_t$. 
In this work, we focus on the throughput of the user falling below a given threshold at time $t$ (see Figure \ref{fig:drive-testing}).\\[0.5em]
\noindent
\textbf{Root Causes.} Let $\vectb{C}=\{c_1, c_2, \dots, c_K\}$ be a set of potential root causes, where each $c_i \in \cal{C}$ corresponds to a cause or factor that may explain symptoms observed in $\vectb{Y}_t$. 

Thus, misconfiguration of some parameters in $\mathbf{U}$ generates symptoms $s_t$, i.e., a faulty system behavior, observable via $\mathbf{Y}_t$, while root causes $c\in\cal{C}$ represents the latent factors explaining the symptom $s_t$, e.g., what is wrongly configured in $\mathbf{U}$. 
The mapping of symptoms to root causes can be captured through a function $f: \cal{U} \times \cal{Y}\times\cal{S} \to \cal{C}$, which represents the underlying physical or causal relationships between root causes and symptoms given the configurations and the observations, as follows: 
\begin{align}
    c = f(\vectb{U}, \vectb{Y}_t, s_t) + \boldsymbol{\epsilon},
\end{align}
where $\boldsymbol{\epsilon}$ represents noise or random variations due to e.g., measurement errors. In general, $f(\cdot)$ is a non-linear function, \emph{partially unknown}, which needs to be inferred from the data.

\subsection{Root Cause Inference}
The goal of RCA is to infer the root cause $c\in\mathbf{C}$, given the network engineering parameters, $\mathbf{U}$, the observation, $\mathbf{Y}_t$, and the symptom, $s_t$.

This can be formulated as an inverse problem:
\begin{equation}\label{eq:probabilistic-rca}
    \hat{c} = \arg\max_{c\in\vectb{C}} \; p\left( c \mid \vectb{U}, \vectb{Y}_t, s_t \right)
\end{equation}

where $p\left(c \mid \vectb{U}, \vectb{Y}_t, s_t \right)$ is the posterior probability of the root causes given the observations.

This formulation provides a probabilistic framework in which root causes are identified based on their likelihood of explaining the observed symptoms. 
However, modeling the posterior probability defined in Eq. \eqref{eq:probabilistic-rca} is very difficult due to the high dimensionality and intricate dependencies in real network data \cite{sole2017survey}.

To solve this issue, we propose training \emph{reasoning} LLMs to provide not only accurate troubleshooting but also coherent, step-by-step explanations of the root causes established on learned patterns from the data and domain-specific knowledge. This approach improves both the interpretability and operational relevance of RCA outcomes. Enabling such capabilities require training on high-quality datasets grounded in domain and system knowledge. To this end, we have developed and made available to the research community the TeleLogs troubleshooting dataset, which is introduced in the following.

\begin{table}[!t]
\renewcommand{\arraystretch}{1.5} 
\centering
\begin{tabular}{|q{1.65cm}|p{6.45cm}|}
\hline
\textbf{Parameter} & \textbf{Description} \\
\hline
\textbf{gNodeB ID} & A unique identifier of a gNodeB in the network. \\
\hline
\textbf{Cell ID} & A unique identifier of a cell. \\
\hline
\textbf{Location} & The GPS coordinates of the \ac{BS}. \\
\hline
\textbf{Mechanical Azimuth} & The horizontal angle of the antenna's direction, measured clockwise from true north. \\
\hline
\textbf{Mechanical Downtilt} & The vertical angle of the antenna's direction, measured downward from the horizontal plane. \\
\hline
\textbf{Digital Tilt} & The virtual antenna tilt, adjustable using software to optimize coverage and performance. \\
\hline
\textbf{Digital \;\;Azimuth} & The virtual antenna azimuth, adjustable using software to optimize coverage and performance. \\
\hline
\textbf{Beam\;\;\;\; Scenario} & The specific beamforming configuration used by the gNodeB to direct the signal. \\
\hline
\textbf{Height} & The height of the gNodeB antenna in meters. \\
\hline
\textbf{PCI} & Physical Cell ID identifying cells in the network.\\
\hline
\textbf{TxRx Mode} & The gNodeB antenna configuration. \\
\hline
\textbf{Max Transmit Power} & The maximum power level at which the gNodeB can transmit signals, measured in dBm. \\
\hline
\textbf{Antenna Model} & The gNodeB antenna model. \\ 
\hline
\end{tabular}
\caption{Network engineering parameters.}
\label{tab:config-data}
\end{table}

\section{The TeleLogs Troubleshooting Dataset}

TeleLogs is a synthetic dataset constructed simulating a network drive testing environment based on real network engineering parameters. In this setup, a \ac{UE} mounted on a moving vehicle, traverses
a region covered by $M$ \acp{BS}, forming a realistic 5G wireless network (as illustrated in Figure \ref{fig:drive-testing}). The simulation provides full visibility into both the network configuration and user-plane performance, enabling fine-grained analysis of fault scenarios.
By providing ground-truth annotations, rich configuration diversity, and controlled simulation variability, TeleLogs enables systematic evaluation of RCA models under realistic 5G network conditions. Both training and testing data are publicly released to foster reproducibility and accelerate progress on reasoning models for mobile networks.

\subsection{Network engineering parameters}

Table \ref{tab:config-data} summarizes the network engineering parameters $\mathbf{U}$ associated with each \ac{BS} (i.e., gNodeB) in the TeleLogs dataset. These include physical deployment attributes (e.g., location, height, cell ID), antenna configurations (e.g., azimuth, downtilt, transmit/receive mode), and beamforming scenarios. Together, these parameters offer a comprehensive view of the network topology and configuration, which are essential for RCA.
\begin{table*}[!t]
\renewcommand{\arraystretch}{1.5} 
\centering
\begin{tabular}{|q{7cm}|p{10cm}|}
\hline
{\textbf{Parameter}} & {\textbf{Description}} \\
\hline
\textbf{Timestamp} & The timestamp indicating when the measurement was recorded. \\
\hline
\textbf{Location (Longitude, Latitude)} & The geographical GPS coordinate of the gNodeB. \\
\hline
\textbf{GPS Speed (km/h)} & The speed of the device at the time of measurement. \\
\hline
\textbf{5G KPI PCell RF Serving PCI} & The PCI of the serving primary cell (PCell).\\
\hline
\textbf{5G KPI PCell RF Serving SS-RSRP [dBm]} & The Reference Signal Received Power (RSRP) of the PCell. \\
\hline
\textbf{5G KPI PCell RF Serving SS-SINR [dB]} & The Signal-to-Interference-plus-Noise Ratio (SINR) of the PCell. \\
\hline
\textbf{5G KPI PCell Layer2 MAC DL Throughput [Mbps]} & The Layer 2 (L2) Medium Access Control (MAC) DL throughput of the PCell. \\
\hline
\textbf{Measurement PCell Neighbor Cell Top Set (Cell Level) Top $k$ PCI} & The PCI of the top-$k$ ($k=1,\dots,5$) neighbor cell in the PCell neighbor cell list. \\
\hline
\textbf{Measurement PCell Neighbor Cell Top Set (Cell Level) Top $k$ Filtered Tx BRSRP [dBm]} & The filtered Transmission Reference Signal Received Power (Tx BRSRP) of the top-$k$ ($k=1,\dots,5$) neighbor cell, measured in dBm. \\
\hline
\textbf{5G KPI PCell Layer1 DL RB Num} & The number of DL Physical Resource Blocks (PRBs) allocated by the PCell. \\
\hline
\end{tabular}
\caption{User plane data.}
\label{tab:user-plane-data}
\end{table*}

\subsection{User plane data}

Table \ref{tab:user-plane-data} presents the user-plane data $\mathbf{Y}_t$ collected during the drive test. It includes timestamped performance indicators, such as downlink (DL) throughput, \ac{RSRP} and \ac{SINR} of the serving primary cell, \ac{RSRP} of the top (i.e., strongest) neighboring cells, and mobility context (GPS speed). These features capture the dynamic interaction between the \ac{UE} and the network, and serve as key observations for RCA.

\subsection{Observed symptom}

The diagnostic scenarios in TeleLogs are centered around a specific symptom: a significant degradation in downlink throughput. Formally, the symptom $s_t$ is defined as the downlink throughput falling below a threshold of 600~Mbps.
Each RCA instance requires identifying the root cause behind such 
performance drop, which may be attributed to factors such as misconfigured mobility or handover parameters, incorrect antenna alignment, or other deployment-related issues.

\subsection{Root Causes}
TeleLogs comprises $K=8$ possible root causes, each corresponding to a distinct misconfiguration or environmental condition that may explain the symptom $s_t$ observed in $\mathbf{Y}_t$:
\begin{enumerate}
    \item[$c_1$)] Test vehicle speed exceeds 40~km/h, {which affects link quality and increases handover frequency.} %impacting user throughput. 
    \item[$c_2$)] The downtilt angle of the serving cell is too large, causing weak coverage at the far end.
    \item[$c_3$)] The serving cell coverage distance exceeds 1~km, resulting in poor RSRP.
    \item[$c_4$)] Non-colocated co-frequency neighboring cells cause severe interference.
    \item[$c_5$)] Neighbor cell and serving cell have the same \ac{PCI} mod 30. As a result, their reference signals can overlap, leading to interference.
    \item[$c_6$)] Frequent handovers degrading user performance.
    \item[$c_7$)] Misconfigured handover thresholds degrading user performance. 
    \item[$c_8$)] {The average \acp{PRB} of the serving cell allocated to the UE is insufficient to reach the target throughput.} 
\end{enumerate}

\section{{Training} Reasoning Models for Network Troubleshooting}

\begin{figure}[!t]
    \centering
    \begin{tcolorbox}[title=\textbf{Prompt template} $\cal{T}$, colframe=black!80, colback=black!10, fonttitle=\bfseries]
        Analyse the user plane data and signalling plane data of a 5G wireless network in the following drive test, as well as the site engineering parameters (tilt, etc.) and configuration (handover parameters, etc.).\\[0.5em]
        \noindent
        Analyse the reasons for $s_t$. Choose the most likely root cause from the following $k$ reasons and include the root cause number in the final answer, enclosed in \textbackslash boxed\{\}.\\[0.5em]
        \noindent
        [Here are the possible reasons $\mathbf{C}$]\\[0.5em]
        \noindent
        [Here are the network configuration data $\mathbf{U}$]\\[0.5em]%engineering parameters
        \noindent
        [Here are the signalling plane data $\mathbf{Y}_t$]
    \end{tcolorbox}
    \caption{RCA Prompt template.}
    \label{fig:prompt-template}
\end{figure}

To solve the problems in TeleLogs, we design reasoning LLMs that approximate the probability distribution in Eq. \eqref{eq:probabilistic-rca} by leveraging their intrinsic \emph{next-token} prediction capability. Specifically, we train an LLM $\pi_\theta$, parameterized by $\theta$, to identify and explain the root cause responsible for a symptom $s_t$ observed from $\mathbf{Y}_t$ on a network with configuration $\mathbf{U}$. To perform this task, we leverage the TeleLogs train dataset with an underlying data distribution $p_{\cal{D}}$. Each training sample is represented as a pair $(q,c)\sim p_{\cal{D}}$, where $q=\mathcal{T}(\vectb{C},\vectb{U}, \vectb{Y}_t, s_t)$ is a structured prompt, specifically designed using the task-specific template $\mathcal{T}(\cdot)$ illustrated in Figure \ref{fig:prompt-template} and $c$ is the ground-truth root cause associated with the prompt $q$. 

Given the prompt $q$, the LLM generates a reasoning trajectory $\tau\sim \pi_\theta(\cdot | q)$ that both identifies and explains the hypothesized root cause $\sigma(\tau) \in \cal{C}\cup\emptyset$, where $\sigma(\cdot)$ is a function mapping trajectories to root causes\footnote{In general, $\sigma(\cdot)$ is a parsing function extracting the final answer enclosed within a \textbackslash boxed\{\} tag that the LLM is prompted to provide.}. This setting makes it possible to define a reward function:
\begin{align}\label{eq:reward-model}
    R(\tau, c) = \mathbbm{1}(\sigma(\tau) = c), 
\end{align}
which provides a binary signal indicating whether the predicted root cause matches the ground truth $c$. Hence, our goal is to learn a policy $\pi_\theta$ that minimizes the following objective function:
\begin{align}\label{eq:objective-function}
    \mathcal{J}(\theta) = &\mathbb{E}_{(q,c)\sim p_\mathcal{D}}\big[\mathbb{E}_{\tau\sim\pi_{\theta(\cdot|q)}}\left[R(\tau, c)\right] - \beta\cal{R}(\theta),
\end{align}

where $\cal{R}(\theta)$ is a term that regularizes the loss function with weight $\beta$. Typically, it is computed as the \ac{KL} divergence \cite{ouyang2022training} between $\pi_\theta$ and a reference policy $\pi_{\rm ref}$: $\cal{R}(\theta)=\mathbb{D}_{\rm KL}(\pi_\theta(\cdot|q), \pi_{\rm ref}(\cdot|q))$.

To solve Eq. \eqref{eq:objective-function}, we use a two-stage training methodology, combining \ac{SFT} and \ac{RL}.
Specifically, we start from an \ac{LLM} with a base policy $\pi_{\theta_0}$ and apply \ac{SFT} to devise a policy $\pi_{\theta_1}$, which in turn undergoes an \ac{RL} training stage to obtain the final policy $\pi_{\theta_2}$.

\begin{figure}
    \centering
    \includegraphics[width=\linewidth]{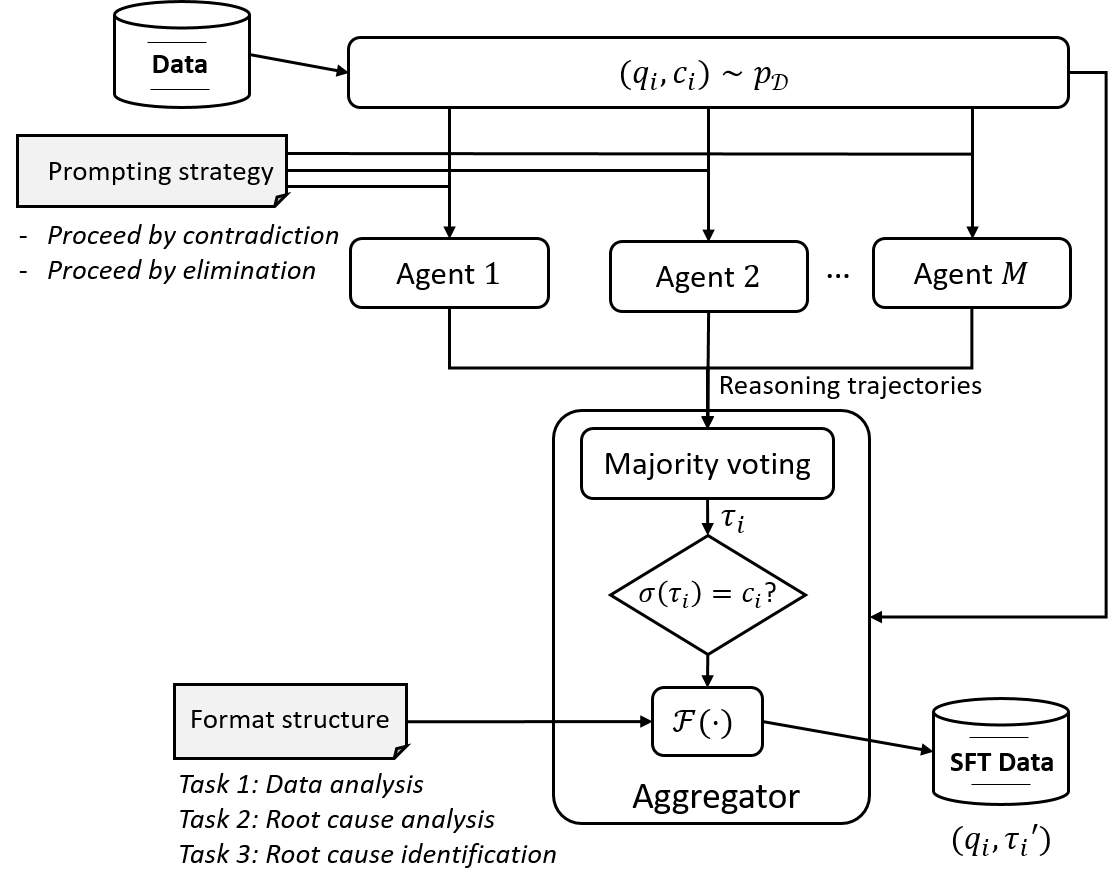}
    \caption{SFT Data generation pipeline.}
    \label{fig:datagen}
\end{figure}

\begin{figure}[!t]
    \centering
    \begin{tcolorbox}[title=\textbf{RCA reasoning format} $\cal{F}$, colframe=black!80, colback=black!10, fonttitle=\bfseries]
        \noindent
        \textbf{Task 1: Data analysis}\\[0.1em]
        We instruct the aggregator to synthesize the data analysis performed by the agents.\\[0.5em]
        \noindent
        \textbf{Task 2: Root cause analysis}\\[0.1em]
        Here, we instruct the aggregator to synthesize the step-by-step reasoning process of the agents, defining the potential root cause(s) based on the analysis made in the previous step.\\[0.5em]
        \noindent
        \textbf{Task 3: Root cause identification}\\[0.1em]
        Here, we instruct aggregator to select the most plausible root cause explaining the observed symptom. \\[0.5em]
        \noindent
        \textbf{Summary:} Eventually, we instruct the aggregator to summarize the RCA and provide the final answer.
    \end{tcolorbox}
    \caption{RCA Reasoning Format.}
    \label{fig:format-template}
\end{figure}

\subsection{Supervised Fine-Tuning}

Fine-tuning with SFT before applying RL offers several compelling benefits when training LLMs \cite{chu2025sft}. First, it gives the model a solid foundation by aligning its outputs with high-quality labeled examples, narrowing down the output distribution to reasonable responses. Second, it improves sample efficiency by teaching the model to produce good outputs, reducing the action-size exploration for the RL and requiring fewer samples and less compute to refine the model further. Eventually, it makes RL training more stable \cite{chu2025sft}.

To be effective, SFT requires high-quality data with explicit reasoning traces demonstrating how to solve problems in the TeleLogs dataset \cite{chu2025sft}. To this end, we design a multi-agent data generation pipeline that produces rich and diverse training samples, ensuring both correctness and reasoning depth.

Our approach leverages multiple LLM-based reasoning agents to perform RCA on observed network issues (see Figure \ref{fig:datagen}). Given an input prompt $q_i$, we instruct a set of $M$ agents to independently perform RCA using carefully designed prompting strategies. Each agent follows a step-by-step reasoning process adopting one of the following tailored strategies:
\begin{itemize}
    \item \textbf{Elimination-based prompting}: the agent systematically evaluates each candidate root cause against the observed symptom and rules out implausible ones.
    \item \textbf{Contradiction-based prompting}: the agent sequentially assumes each candidate root cause to be the most plausible one, thereby evaluating its implications against the observed data. If the assumption leads to a contradiction, it is discarded, and the process continues until a consistent root cause is identified.
\end{itemize}

\begin{figure}[!t]
    \centering
    \includegraphics[width=\linewidth]{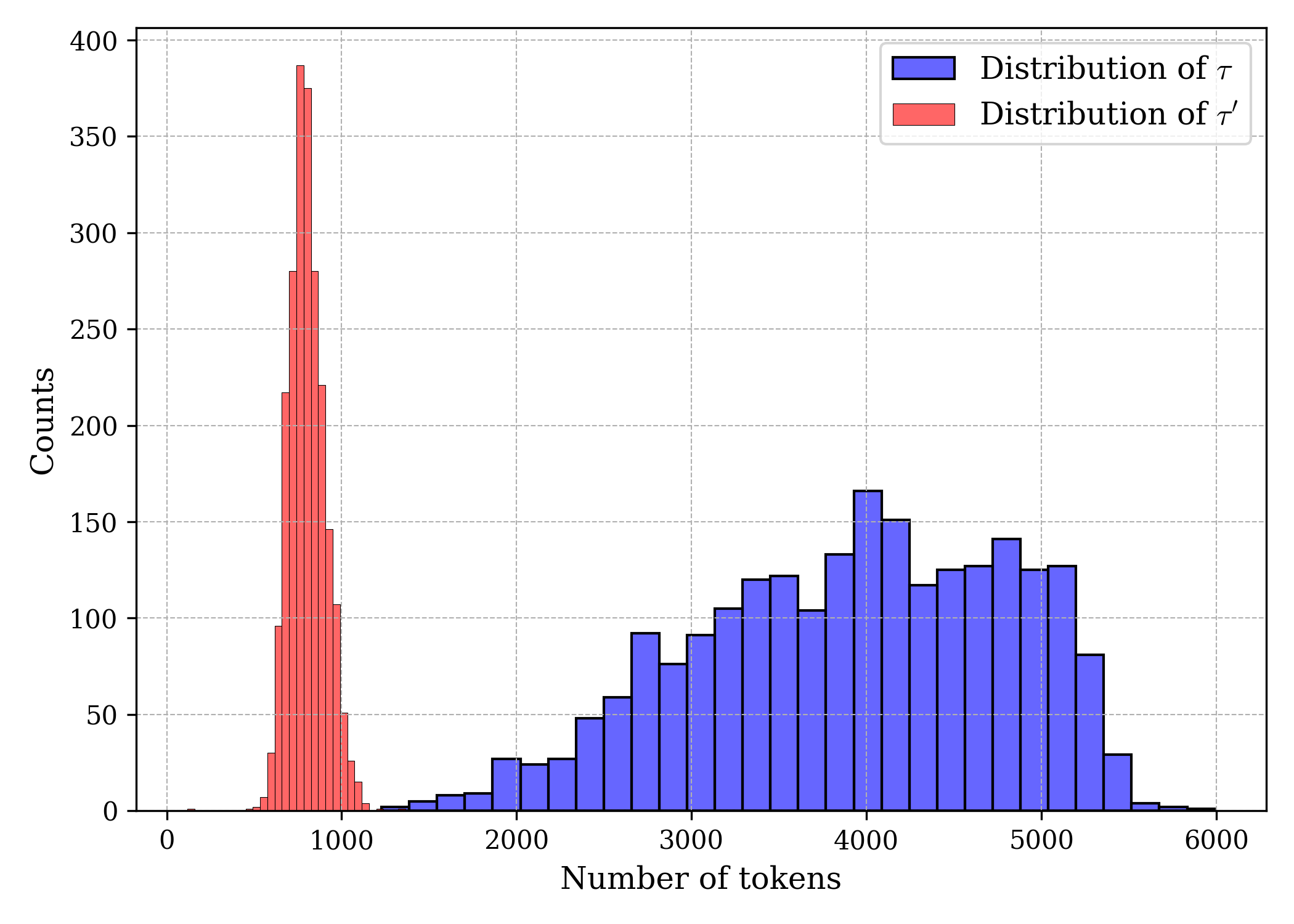}
    \caption{Data distribution before and after the aggregator.}
    \label{fig:data-dist}
\end{figure}

By integrating different selection strategies, our approach allows for diverse and richer diagnostic trajectories across agents, potentially enriching the solutions with plausible perspectives. However, although reasoning trajectories provide a deep understanding on the reflections made by the agents to select the root causes, they often include redundant steps such as backtracking or repetitive verification, which can obscure the core rationale \cite{chen2024not}. To improve interpretability and usability in real networks, we introduce an \emph{aggregator} agent designed to synthesize a concise and structured explanation of the root cause decision.
The aggregator agent first selects the final solution trajectory $\tau_i$ based on majority voting across the $M$ candidate trajectories $(\tau_k,;k=1,\dots,M)$. If the predicted root cause $\sigma(\tau_i)$ matches the ground truth $c_i$, the aggregator reformulates the trajectory into a structured and compact format using the template $\mathcal{F}$, as illustrated in Figure~\ref{fig:format-template}. This step produces the final RCA trace $\tau_i'=\cal{F}(\tau_i)$ with significantly reduced number of tokens, i.e. $|\tau_i'| \ll |\tau_i|$ (see Figure \ref{fig:data-dist}), thereby improving the sample efficiency of the SFT model.
Hence, we construct a solid training data $\cal{D}'=\{(q_i, \tau_i')\}_i$ with an underlying distribution $p_{\cal{D}'}$, keeping only the good trajectories with $\sigma(\tau_i)=c_i$, which we use for SFT to get $\pi_{\theta_1}$ by minimizing a token-level cross-entropy loss:
\begin{align}\label{eq:sft-training}
    \cal{J}_{\rm SFT}(\theta) = -\mathbb{E}_{(q, \tau')\sim p_{\cal{D}'}}\left[\frac{1}{|\tau'|} \sum_{j=1}^{|\tau'|} \mathrm{log}\ \pi_{\theta}(\tau_{j}' |q, \tau_{<j}')\right].
\end{align}
In Eq. \eqref{eq:sft-training}, $\tau_{j}'$ denotes the $j$-th token of $\tau'$, $\tau_{<j}'$ is the sequence of tokens preceding the $j$-th token, and $|\tau'|$ denotes the number of tokens of $\tau'$.

\subsection{RL Fine-Tuning}
The fine-tuned SFT policy $\pi_{\theta_1}$ undergoes RL training using GRPO to get $\pi_{\theta_2}$. In GRPO, the first term of the objective function Eq. \eqref{eq:objective-function} is replaced with 
\begin{align}
    \mathcal{J}_{\rm RL}(\theta) = \mathbb{E}_{\substackleft{(q,\cdot)\sim p_\cal{D},\\ \{\tau_i\}_{i=1}^N\sim \pi_{\rm old}(\cdot|q)}}\left[\frac{1}{N} \sum_{i=1}^N \frac{1}{|\tau_i|} \sum_{j=1}^{|\tau_i|} \rho_{i,j}(\theta)\right],
    %- \beta\cal{R}(\theta)
\end{align}
where 
\[
\rho_{i,j}(\theta) = \min\left[\eta_{i,j}(\theta) \hat{A}_{i,j}, \mathrm{clip}(\eta_{i,j}, 1-\epsilon, 1+\epsilon)\hat{A}_{i,j} \right],
\] 
 $\eta_{i,j}(\theta) = \frac{\pi_\theta(\tau_{i,j} | q, \tau_{i,< j})}{\pi_{\rm old}(\tau_{i,j} | q, \tau_{i,< j})}$ is the probability ratio between the old and current policy, $\hat{A}_{i,j}$ is the advantage estimate of the $j$-th token $\tau_{i,j}$ and $\epsilon$ is the clipping constant, which prevents excessively large policy updates \cite{schulman2017ppo}.

In GRPO, the advantage is estimated on the basis of the group response--the rewards of $N$ trajectories sampled for each question:
\begin{equation}
    \hat{A}_{i,j} = \frac{r_{i,j} - \text{mean}(\mathbf{r}_j)}{\text{std}(\mathbf{r}_j)}
\end{equation}
where $\mathbf{r}_j = \{r_{1,j}, r_{2,j}, \dots, r_{N,j}\}$ and $r_{i,j}$ is the token-level reward. In our setting, all tokens receive equal reward corresponding to the one computed for the entire trajectory using Eq. \eqref{eq:reward-model}. However, the effective reward per token comprises its contribution in the \ac{KL} loss $\cal{R}(\theta)$ \cite{sheng2024hybridflow}.

\section{Results}

In this section, we present a comprehensive evaluation of our proposed training methodology for RCA using the TeleLogs dataset. Our experiments focus on models from the Qwen family, trained at three different scales: Qwen2.5-1.5B-instruct, Qwen2.5-7B-instruct, and Qwen2.5-32B-instruct \cite{yang2025qwen3}. Each model is fine-tuned for 10 epochs, totaling 1500 training steps, with a batch size of 128 using the VERL framework \cite{sheng2024hybridflow}. 
To generate high-quality SFT training data, we employ a multi-agent pipeline (Figure~\ref{fig:datagen}) involving $M=2$ reasoning agents based on Qwen3 32B and QwQ 32B models \cite{yang2025qwen3}.
During training, we sample $N=8$ trajectories per question to encourage diverse reasoning paths. We empirically set $\epsilon=0.2$ and use a learning rate of $10^{-6}$.

We evaluate the performance of models on the TeleLogs test set by generating $N=4$ response samples for each test instance. To assess accuracy and reasoning consistency, we report two complementary evaluation metrics:
\begin{itemize}
    \item \textbf{pass@1}: This metric measures the ability of the model to produce a correct answer in a single attempt. It is computed by evaluating each of the $N$ generated responses individually and averaging the correctness over all samples.
    \item \textbf{maj@4}: This metric captures the model’s consistency and reliability across multiple trajectories. It performs majority voting over the $N=4$ generated answers for each question. A test instance is considered correctly solved if the most frequently occurring answer among the four is correct.
\end{itemize}

\begin{figure}[!t]
    \centering
    \includegraphics[width=\linewidth]{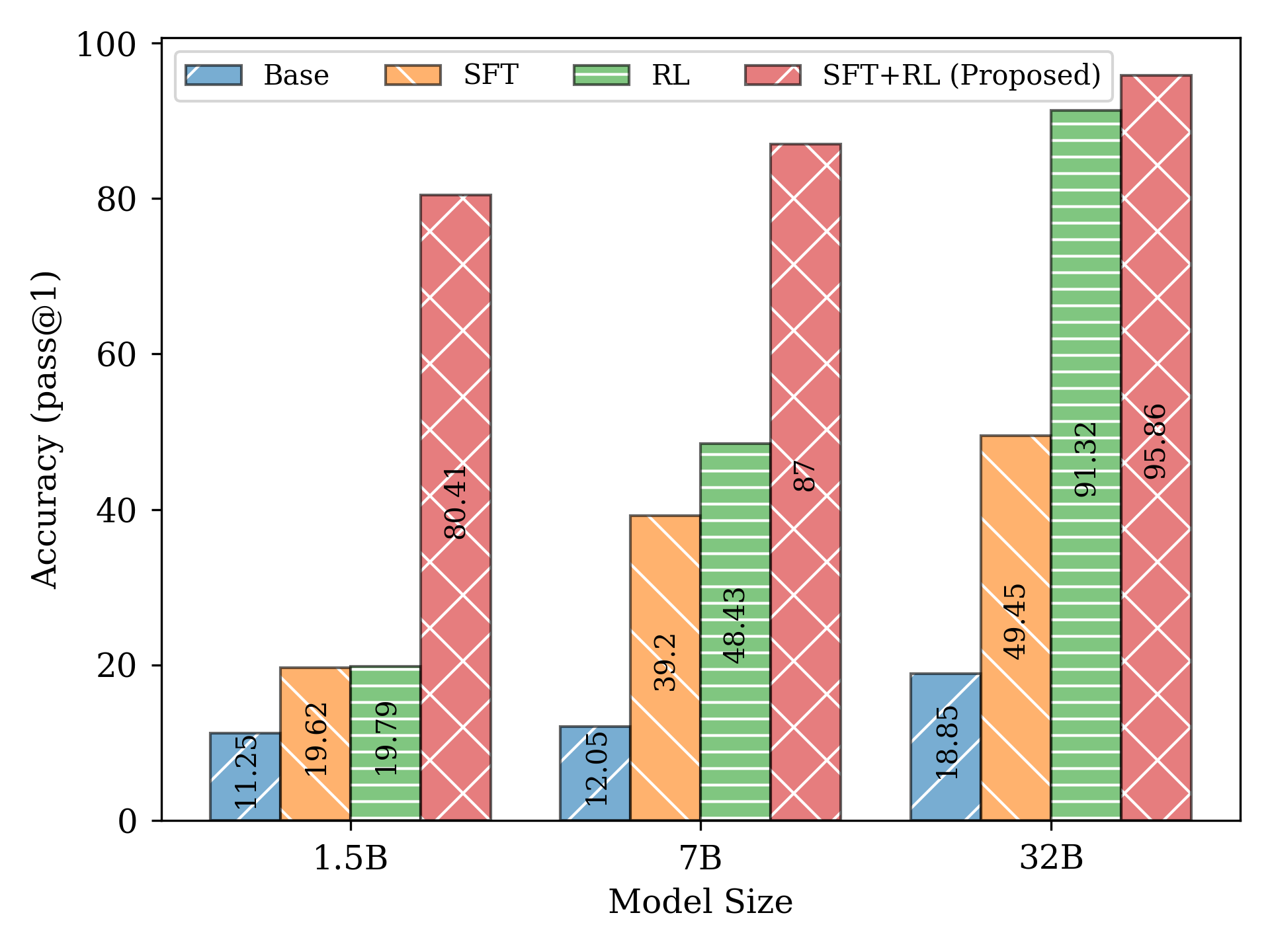}
    \caption{Performance comparison of different training methodologies.}
    \label{fig:test-results}
\end{figure}

\subsection{Performance of different training methodologies}
\label{sec:1st_results}

Figure~\ref{fig:test-results} presents the performance of our proposed training methodology (SFT+RL) in comparison to three baselines: the untrained base model (Base), SFT alone, and RL alone, across three different model sizes: 1.5B, 7B, and 32B.

Across all model sizes, we observe a consistent and significant improvement when using our proposed training methodology. For the smallest model, Qwen2.5-RCA-1.5B, our method achieves an accuracy of $80.41\%$, outperforming both SFT ($19.6\%$) and RL ($19.79\%$) by a large margin, and showing a 7× gain over the base model ($11.25\%$). This illustrates the advantage of combining SFT and RL for relatively small models.

As the model size increases, performance improves across all methods, but the gains are particularly pronounced with SFT+RL. At the 7B scale, the proposed method achieves $87\%$ accuracy, compared to $39.2\%$ for RL and $48.43\%$ for SFT. Notably, while RL alone provides better gains than SFT for 7B, the combined method substantially exceeds both.

For the largest model, Qwen2.5-32B, training directly using RL is more effective compared to 1.5B and 7B models. However, our approach still exhibits an accuracy of $95.86\%$, compared to $91.32\%$ (RL), $49.45\%$ (SFT), and $18.5\%$ for the base model. These results demonstrate that \ac{RL} more effectively refines the reasoning and decision-making abilities of LLMs, especially medium size models, when it is preceded by \ac{SFT}.

\begin{table}[!t]
\centering
\begin{tabular}{zcccc}
\rowcolor{white}
& & \multicolumn{2}{c}{\textbf{Accuracy} (\%)} &   \\
\cmidrule(lr){2-5}
&\multicolumn{2}{c}{\textbf{Test dataset}} & \multicolumn{2}{c}{\textbf{Randomized dataset}}  \\
\cmidrule(lr){2-3}
\cmidrule(lr){4-5}
\rowcolor{white}
\textbf{Ref. Model} & \textbf{Pass@1} &\textbf{Maj@4} & \textbf{Pass@1} &\textbf{Maj@4}\\
\midrule
{\textbf{Base}} & & & & \\
\cmidrule(lr){0-0}
\shortstack{Qwen2.5\\1.5B-Instruct} & {11.25} & {11.6}& {9.15}& {9.8}\\[1em]
\shortstack{Qwen2.5\\7B-Instruct} & {12.05}& {10.8} & {11.5}& {11.8} \\[1em]
\shortstack{Qwen2.5\\32B-Instruct} & {18.85}& {19.6} & {18.05}& {18.7} \\
\midrule
{\textbf{Reasoning}} & & & & \\
\cmidrule(lr){0-0}

\shortstack{{DeepSeek-R1}\\{Distill-Llama-70B}} & 29.42 & 34.84 & 29 & 32.18  \\[1em]
\shortstack{QwQ\\32B} & 33.62 & 39 & 32.14 & 38.86 \\[1em] %
 \shortstack{Qwen3\\32B} & 33.77 & 37.04 & 31.37 & 36.23 \\
\midrule

{\textbf{Proposed}} & & & & \\
\cmidrule(lr){0-0}
\rowcolor{gray!10}
 \shortstack{Qwen2.5-RCA\\1.5B} & {87.56} & {87.73}& {75.90}& {77.08}\\[1em]
\rowcolor{gray!10}
\shortstack{Qwen2.5-RCA\\7B} & {87.01}& {88.89}& {77.95}& {80.32}\\[1em]
\rowcolor{gray!10}
\shortstack{Qwen2.5-RCA\\32B} & \textbf{95.86}& \textbf{96.18} & \textbf{93.23}& \textbf{95.02}\\
  
\bottomrule
\end{tabular}
\caption{Performance comparison.}
\label{tab:perf-comparison}
\end{table}

\subsection{Performance comparison: Fine-tuned Small models outperform SoTA LLMs} %

In this section, we extend the results presented in Sec. \ref{sec:1st_results} by comparing the performance of our fine-tuned models against SotA base and reasoning LLMs of different sizes.

As shown in Table~\ref{tab:perf-comparison}, our fine-tuned models significantly outperform both base and reasoning LLMs. For instance, Qwen2.5-RCA-32B achieves a \texttt{pass@1} of $95.86\%$ and maj@4 of $96.18\%$, far surpassing strong baselines such as Qwen3-32B ($33.77\%$, $34.04\%$) and DeepSeek R1 Distill-Llama-70B ($29.42\%$, $34.84\%$).

Even at smaller scales, our fine-tuned models show dramatic gains. For example, Qwen2.5-RCA-1.5B reaches $87.56\%$ \texttt{pass@1}, which is over 2.5× higher than the SoTA reasoning models even the DeepSeek R1 Distill-Llama-70B. This demonstrates the effectiveness of our two-stage training approach in enabling reasoning LLMs to correctly identify and explain root causes in complex network scenarios. Figure \ref{fig:rca-traces} shows an example of RCA traces using our Qwen2.5-RCA-32B model.

\subsection{Generalization to the Randomized Test Dataset}

To evaluate robustness and generalization of the proposed methodology, we further test the LLMs on a randomized version of the dataset, where the root cause identifiers, table order, and other superficial cues are altered. This setup is designed to prevent models from relying on position-based heuristics or memorized patterns. Our fine-tuned models continue to show strong performance under this randomized setting. Qwen2.5-32B retains high accuracy, with $93.23\%$ \texttt{pass@1} and $95.02\%$ maj@4. Smaller models trained with our methodologies suffer of larger performance drop; however, the 1.5B and 7B models maintain performance above $75\%$, indicating that the models are not simply overfitting to surface-level patterns but are learning robust causal reasoning strategies.

\section{Conclusion}

We introduced a lightweight yet effective framework for Root Cause Analysis (RCA) using reasoning-capable Large Language Models (LLMs), supported by a new synthetic dataset, TeleLogs, which captures realistic 5G troubleshooting scenarios. Our approach combines supervised fine-tuning with reinforcement learning to improve both the accuracy and the interpretability of RCA predictions.

Experiments across multiple model scales show that our method outperforms both instruction-tuned and state-of-the-art reasoning LLMs, achieving over $95\%$ accuracy with Qwen2.5-32B-Instruct on the TeleLogs test set. The approach also generalizes well to randomized variants, confirming its robustness. These results suggest that reasoning LLMs, when properly trained, can serve as powerful and explainable diagnostic tools for complex systems. Future work will focus on extending the method to handle multiple root cause scenarios and incorporating real-world operational data.

\section*{Appendix}
\label{appendix}
In this section, we show an example of RCA (see Figure \ref{fig:rca-traces}) on a problem from TeleLogs (see Figure \ref{fig:rca-problem}).

\section*{Author Statement}
We open source the TeleLogs dataset under MIT Licence to support research on RCA in 5G networks. All data are anonymized and aggregated to ensure privacy and confidentiality.

\begin{figure*}[!t]
    \centering
    \resizebox{\textwidth}{!}{% 
    \begin{tcolorbox}[colback=gray!5, colframe=black, title=Root Cause Reasoning Trace  (Qwen2.5-RCA-32B | \textbf{Thought for 11 seconds}), fonttitle=\bfseries]
    \small
    Okay, let's tackle this problem step by step. First, I need to understand the data provided and find out why the user throughput dropped.
    
    \vspace{0.5em}
    \textbf{Task 1: Data Analysis}
    \begin{itemize}
      \item \textbf{Throughput Drop:} The throughput drops to \textbf{0.14 Mbps} and \textbf{13.23 Mbps} when the serving cell is \textbf{PCI 919}. After switching to \textbf{PCI 737}, it recovers to \textbf{346.52 Mbps} and above.
      \item \textbf{Neighbor Cells:} PCI 737 (a top neighbor of PCI 919) becomes the serving cell at 10:25:37, coinciding with throughput recovery. PCI 737’s throughput is consistently high when it is the serving cell.
      \item \textbf{PCI Modulo 30 Check:} PCI 919 mod 30 = 19; PCI 737 mod 30 = 17. \textbf{No conflict (C6 unlikely)}.
      \item \textbf{Downtilt/Azimuth:} PCI 919’s mechanical downtilt is 4°, digital tilt totals 8°, and azimuth is 100°, potentially limiting coverage in some directions.
      \item \textbf{Coverage Distance:} Serving cell distance is minimal $<100$m. (\textbf{C2 unlikely}).
      \item \textbf{Overlapping Coverage:} PCI 737 and PCI 919 belong to the same gNodeB (0000258), so non-colocated interference (C4) is less likely.
      \item \textbf{Frequent Handover:} Only one handover to PCI 737. \textbf{C5 unlikely}.
      \item \textbf{RBs:} RB values (e.g., 160–186) are high. \textbf{C8 unlikely}.
      \item \textbf{Test Speed:} Low speeds ($\leq 34$ km/h). \textbf{C7 unlikely}.
    \end{itemize}
    
    \vspace{0.5em}
    \textbf{Task 2: Root Cause Analysis}
    \begin{enumerate}
      \item \textbf{C1 (Excessive Downtilt):} PCI 919’s total downtilt is 12° (mechanical 4° + digital 8°), but its Beam Scenario (SCENARIO\_1) has a narrow vertical beamwidth (6°). This may cause weak coverage at the far end, but the test points are very close ($\sim 5-20$m).
      \item \textbf{C2 (Over-Shooting Coverage):} Serving distance is $<100$m (within 1 km), so this is invalid.
      \item \textbf{C3 (Higher Neighbor Throughput):} PCI 737 provides \textbf{much higher throughput}. The test device should have been served by PCI 737 earlier to avoid the drop.
      \item \textbf{C4 (Overlapping Coverage):} PCI 737 and PCI 919 are colocated (same gNodeB), so non-colocated interference is ruled out.
      \item \textbf{C5 (Frequent Handovers):} Only one handover occurred.
      \item \textbf{C6 (PCI Mod 30 Conflict):} No PCI conflict detected.
      \item \textbf{C7 (High Test Speed):} Speed is low.
      \item \textbf{C8 (Low RBs):} RBs are sufficient.
    \end{enumerate}
    
    \vspace{0.5em}
    \textbf{Conclusion:} The throughput drop occurs because the test device remains in the weak coverage area of PCI 919 until a handover to PCI 737 (which has ~10× higher throughput ) occurs. PCI 737 should have been the serving cell earlier to avoid the drop.
    
    \vspace{0.5em}
    \textbf{Task 3: Most Likely Root Cause}
    
    The root cause is \textbf{C3}: The neighbor cell (PCI 737) provides significantly higher throughput. The serving cell (PCI 919) should have triggered a handover to PCI 737 earlier to prevent the throughput degradation.
    
    \[
    \boxed{\text{C3}}
    \]
    
    \vspace{0.5em}
    \textbf{Summary:}
    The root cause of the throughput degradation is identified as C3: the test device experiences a significant drop in throughput while connected to the serving cell PCI 919, which has lower throughput capacity compared to the neighboring cell PCI 737. The throughput recovers once the device is handed over to PCI 737, which provides approximately 10 times higher throughput. This indicates that an earlier handover to PCI 737 would have prevented the throughput drop.
    \end{tcolorbox}
    }
    \caption{RCA traces. It includes data interpretation, elimination of unlikely causes, and a final conclusion pointing to the most probable root cause, with justification. The reasoning follows a structured analysis based on network KPIs, configuration parameters, and RCA criteria.}
    \label{fig:rca-traces}
\end{figure*}

\begin{figure*}[!b]
    \centering
    \begin{tcolorbox}[colback=gray!5, colframe=black, title=RCA Prompt (5G Troubleshooting), fonttitle=\bfseries]
    \small
    \textbf{Analyze the 5G wireless network drive-test user plane data and engineering parameters.} Identify the reason for the throughput dropping below \textbf{600 Mbps} in certain road sections. From the following 8 potential root causes, select the most likely one and enclose its number in \verb|\boxed{}| in the final answer.
    
    \vspace{0.5em}
    
    \textbf{C1}: The serving cell's downtilt angle is too large, causing weak coverage at the far end. \\
    \textbf{C2}: The serving cell's coverage distance exceeds 1 km, resulting in over-shooting. \\
    \textbf{C3}: A neighboring cell provides higher throughput. \\
    \textbf{C4}: Non-colocated co-frequency neighboring cells cause severe overlapping coverage. \\
    \textbf{C5}: Frequent handovers degrade performance. \\
    \textbf{C6}: Neighbor cell and serving cell have the same PCI mod 30, leading to interference. \\
    \textbf{C7}: Test vehicle speed exceeds 40 km/h, impacting user throughput. \\
    \textbf{C8}: Average scheduled RBs are below 160, affecting throughput.
    
    \vspace{0.5em}
    
    \textbf{Given:} \\
    - The default electronic downtilt value is 255, representing a downtilt angle of 6 degrees. Other values represent the actual downtilt angle in degrees.
    
    \vspace{0.5em}
    \textbf{Beam Scenario and Vertical Beamwidth Relationships:}
    \begin{itemize}
      \item When the cell's Beam Scenario is set to \texttt{Default} or \texttt{SCENARIO\_1} to \texttt{SCENARIO\_5}, the vertical beamwidth is 6 degrees.
      \item When the cell's Beam Scenario is set to \texttt{SCENARIO\_6} to \texttt{SCENARIO\_11}, the vertical beamwidth is 12 degrees.
      \item When the cell's Beam Scenario is set to \texttt{SCENARIO\_12} or above, the vertical beamwidth is 25 degrees.
    \end{itemize}
    
    \vspace{0.5em}
    \textbf{User plane drive test data as follows:}
    \begin{lstlisting}[basicstyle=\ttfamily\scriptsize, frame=single,breakautoindent=false,breakindent=0ex, breaklines=true]
    Timestamp|Longitude|Latitude|GPS Speed (km/h)|5G KPI PCell RF Serving PCI|5G KPI PCell RF Serving SS-RSRP [dBm]|5G KPI PCell RF Serving SS-SINR [dB]|5G KPI PCell Layer2 MAC DL Throughput [Mbps]|Measurement PCell Neighbor Cell Top Set(Cell Level) Top 1 PCI|...|5G KPI PCell Layer1 DL RB Num (Including 0)
    2025-05-07 10:25:34|128.139682|32.623035|34|919|-80.48|11.59|600.0|737|...|161.0
    2025-05-07 10:25:35|128.139717|32.622993|28|919|-78.15|8.5|0.14|737|...|160.0
    2025-05-07 10:25:36|128.139745|32.622954|1|919|-82.19|8.41|13.23|737|...|186.0
    2025-05-07 10:25:37|128.139781|32.622904|38|737|-88.07|11.04|346.52|919|...|180.0
    2025-05-07 10:25:38|128.139809|32.622862|32|737|-78.39|17.76|515.45|919|...|173.2
    2025-05-07 10:25:39|128.139837|32.622823|0|737|-77.94|15.01|1056.42|919|...|168.69
    2025-05-07 10:25:40|128.139872|32.622781|6|737|-78.33|14.93|1085.04|919|...|165.05
    2025-05-07 10:25:41|128.1399|32.622743|22|737|-83.87|10.86|1102.15|919|...|161.93
    2025-05-07 10:25:42|128.139929|32.6227|29|737|-81.79|11.52|1091.58|919|...|171.97
    2025-05-07 10:25:43|128.139964|32.622662|7|737|-84.77|7.05|1010.69|919|...|177.81
    \end{lstlisting}
    
    \vspace{0.5em}
    \textbf{Engineering parameters data as follows:}
    \begin{lstlisting}[basicstyle=\ttfamily\scriptsize, frame=single, breakautoindent=false,breakindent=0ex, breaklines=true]
    gNodeB ID|Cell ID|Longitude|Latitude|Mechanical Azimuth|Mechanical Downtilt|Digital Tilt|Digital Azimuth|Beam Scenario|Height|PCI|TxRx Mode|Max Transmit Power|Antenna Model
    0000258|1|128.139529|32.623035|45|3|7|5|SCENARIO_7|9.0|737|32T32R|34.9|NR AAU 1
    0000258|26|128.139529|32.623035|145|6|255|0|DEFAULT|9.0|291|32T32R|34.9|NR AAU 1
    0000258|15|128.139529|32.623042|100|4|8|0|SCENARIO_1|15.0|919|32T32R|34.9|NR AAU 1
    0000258|5|128.14087|32.621659|310|5|255|0|DEFAULT|14.7|430|32T32R|34.9|NR AAU 1
    0000258|24|128.140904|32.621691|55|0|6|0|DEFAULT|14.7|420|32T32R|34.9|NR AAU 1
    0000570|16|128.144983|32.619395|20|10|255|0|DEFAULT|90.0|36|64T64R|34.9|NR AAU 2
    \end{lstlisting}
    \end{tcolorbox}
    \caption{An example of RCA Problem. It includes a description of the RCA task, a list of potential root causes (C1-C8), domain-specific rules regarding downtilt and beam scenario relationships, and structured input data comprising user-plane drive-test logs and engineering parameters. The problem description provides all necessary context for the model to reason.}
    \label{fig:rca-problem}
\end{figure*}

\begin{acronym}[AAAAAAAAA]
  \acro{3GPP}{Third Generation Partnership Project}
  \acro{QnA}{Question and Answer}
 \acro{AI} {Artificial Intelligence}
 \acro{BERT}{Bidirectional Encoder Representations from Transformers}
 \acro{CoT}{chain-of-thought}
  \acro{DL}{Deep Learning}
   \acro{FPGA}{Field-Programmable Gate Array}
  \acro{GPT}{Generative Pre-trained Transformer}
  \acro{GNN}{graph neural networks}
  \acro{GRPO}{group relative policy optimization}
  \acro{KPI}{Key Performance Indicator}
 \acro{LLM}{large language model}
  \acro{MNO}{Mobile Network Operator}
\acro{BS}{Base Station}
  \acro{ML}{Machine Learning}
 \acro{NLP}{Natural Language Processing} 
 \acro{OandM}[O\&M]{operation and management}
 \acro{API}{Application Programming Interface} 
 \acro{PCI}{physical cell ID }
 \acro{RAN}{Radio Access Network}
 \acro{RAG}{Retrieval Augmented Generation}
 \acro{RB}{resource block}
 \acro{PRB}{physical resource block}
 \acro{RCA}{root cause analysis}
 \acro{RSRP}{reference signal received power}
 \acro{SINR}{signal-to-interference-plus-noise ratio}
 \acro{UE}{user equipment}
 \acro{MIMO}{multiple-input multiple-output}
 \acro{RL}{Reinforcement Learning}
 \acro{SFT}{Supervised Fine-Tuning}
 \acro{PPO}{Proximal Policy Optimization}
 \acro{KL}{Kullback-Leibler}
 \end{acronym}

\clearpage
\clearpage
\bibliographystyle{ieeetr}
\bibliography{bibliography}

\begin{thebibliography}{10}

\bibitem{zhang2022icassp}
T.~Zhang, Q.~Chen, Y.~Jiang, D.~Miao, F.~Yin, T.~Quan, Q.~Shi, and Z.-Q. Luo, ``{ICASSP-SPGC 2022: Root cause analysis for wireless network fault localization},'' in {\em IEEE International Conference on Acoustics, Speech and Signal Processing (ICASSP)}, pp.~9301--9305, IEEE, 2022.

\bibitem{tong2021machine}
V.~Tong, S.~Souihi, H.~A. Tran, and A.~Mellouk, ``Machine learning based root cause analysis for sdn network,'' in {\em 2021 IEEE Global Communications Conference (GLOBECOM)}, pp.~1--6, IEEE, 2021.

\bibitem{gonzalez2017root}
J.~M.~N. Gonzalez, J.~A. Jimenez, J.~C.~D. Lopez, {\em et~al.}, ``Root cause analysis of network failures using machine learning and summarization techniques,'' {\em IEEE Communications Magazine}, vol.~55, no.~9, pp.~126--131, 2017.

\bibitem{sole2017survey}
M.~Sol{\'e}, V.~Munt{\'e}s-Mulero, A.~I. Rana, and G.~Estrada, ``Survey on models and techniques for root-cause analysis,'' {\em arXiv preprint arXiv:1701.08546}, 2017.

\bibitem{Yen22GNN}
C.-C. Yen, W.~Sun, H.~Purmehdi, W.~Park, K.~R. Deshmukh, N.~Thakrar, O.~Nassef, and A.~Jacobs, ``Graph neural network based root cause analysis using multivariate time-series kpis for wireless networks,'' in {\em IEEE/IFIP Network Operations and Management Symposium}, pp.~1--7, 2022.

\bibitem{mata24SpatialGraph}
L.~Mata, M.~Sousa, P.~Vieira, M.~P. Queluz, and A.~Rodrigues, ``On the use of spatial graphs for performance degradation root-cause analysis toward self-healing mobile networks,'' {\em IEEE Access}, vol.~12, pp.~20490--20508, 2024.

\bibitem{tnAutoRCA2025}
K.~Wu, Q.~Yu, M.~Mei, R.~Liu, J.~Wang, K.~Zhang, Y.~Bao, R.~Ye, B.~He, J.~Liao, L.~Huang, Y.~Du, Z.~Yang, K.~Liu, Z.~Song, Y.~Gao, F.~Tan, J.~Yang, and N.~Gu, ``{TN-AutoRCA: Benchmark Construction and Agentic Framework for Self-Improving Alarm-Based Root Cause Analysis in Telecommunication Networks},'' {\em arXiv preprint arXiv:2507.18190v1}, 2025.

\bibitem{roy2024exploring}
D.~Roy, X.~Zhang, R.~Bhave, C.~Bansal, P.~Las-Casas, R.~Fonseca, and S.~Rajmohan, ``{Exploring llm-based agents for root cause analysis},'' in {\em Companion Proceedings of the 32nd ACM International Conference on the Foundations of Software Engineering}, pp.~208--219, 2024.

\bibitem{wang2024rcagent}
Z.~Wang, Z.~Liu, Y.~Zhang, A.~Zhong, J.~Wang, F.~Yin, L.~Fan, L.~Wu, and Q.~Wen, ``{Rcagent: Cloud root cause analysis by autonomous agents with tool-augmented large language models},'' in {\em Proceedings of the 33rd ACM International Conference on Information and Knowledge Management}, pp.~4966--4974, 2024.

\bibitem{pei2025flow}
C.~Pei, Z.~Wang, F.~Liu, Z.~Li, Y.~Liu, X.~He, R.~Kang, T.~Zhang, J.~Chen, J.~Li, {\em et~al.}, ``{Flow-of-Action: SOP Enhanced LLM-Based Multi-Agent System for Root Cause Analysis},'' in {\em Companion Proceedings of the ACM on Web Conference 2025}, pp.~422--431, 2025.

\bibitem{ouyang2022training}
L.~Ouyang, J.~Wu, X.~Jiang, D.~Almeida, C.~Wainwright, P.~Mishkin, C.~Zhang, S.~Agarwal, K.~Slama, A.~Ray, {\em et~al.}, ``Training language models to follow instructions with human feedback,'' {\em Advances in neural information processing systems}, vol.~35, pp.~27730--27744, 2022.

\bibitem{chu2025sft}
T.~Chu, Y.~Zhai, J.~Yang, S.~Tong, S.~Xie, D.~Schuurmans, Q.~V. Le, S.~Levine, and Y.~Ma, ``Sft memorizes, rl generalizes: A comparative study of foundation model post-training,'' {\em arXiv preprint arXiv:2501.17161}, 2025.

\bibitem{chen2024not}
X.~Chen, J.~Xu, T.~Liang, Z.~He, J.~Pang, D.~Yu, L.~Song, Q.~Liu, M.~Zhou, Z.~Zhang, {\em et~al.}, ``Do not think that much for 2+ 3=? on the overthinking of o1-like llms,'' {\em arXiv preprint arXiv:2412.21187}, 2024.

\bibitem{schulman2017ppo}
J.~Schulman, F.~Wolski, P.~Dhariwal, A.~Radford, and O.~Klimov, ``{Proximal Policy Optimization Algorithms},'' {\em CoRR}, vol.~abs/1707.06347, 2017.

\bibitem{sheng2024hybridflow}
G.~Sheng, C.~Zhang, Z.~Ye, X.~Wu, W.~Zhang, R.~Zhang, Y.~Peng, H.~Lin, and C.~Wu, ``{HybridFlow: A Flexible and Efficient RLHF Framework},'' {\em arXiv preprint arXiv: 2409.19256}, 2024.

\bibitem{yang2025qwen3}
A.~Yang, A.~Li, B.~Yang, B.~Zhang, B.~Hui, B.~Zheng, B.~Yu, C.~Gao, C.~Huang, C.~Lv, {\em et~al.}, ``Qwen3 technical report,'' {\em arXiv preprint arXiv:2505.09388}, 2025.

\end{thebibliography}

\end{document}